\documentclass[10pt,twocolumn,letterpaper]{article}

\usepackage{cvpr}              

\usepackage{graphicx}
\usepackage{amsmath}
\usepackage{amssymb}
\usepackage{booktabs}
\usepackage{multirow}
\usepackage{titling}
\usepackage{enumitem} 
\usepackage[super]{nth}
\usepackage[belowskip=-5pt,aboveskip=2pt]{caption}
\usepackage[pagebackref,breaklinks,colorlinks]{hyperref}

\usepackage[capitalize]{cleveref}
\crefname{section}{Sec.}{Secs.}
\Crefname{section}{Section}{Sections}
\Crefname{table}{Table}{Tables}
\crefname{table}{Tab.}{Tabs.}

\def\cvprPaperID{64} 
\def\confName{CVPR Workshop NTIRE}
\def\confYear{2022}

\begin{document}

\title{{A New Dataset and Transformer for Stereoscopic Video Super-Resolution 
}}

\author{Hassan Imani$^{1}$, Md Baharul Islam$^{1,2}$, Lai-Kuan Wong$^{3}$\\
$^{1}$Bahcesehir University \quad \quad \quad
$^{2}$American University of Malta \quad \quad \quad
$^{3}$Multimedia University\\
{\tt\small hassan.imani1987@gmail.com, bislam.eng@gmail.com, lkwong@mmu.edu.my}
}

\maketitle

\begin{abstract}

Stereo video super-resolution (SVSR) aims to enhance the spatial resolution of the low-resolution video by reconstructing the high-resolution video. The key challenges in SVSR are preserving the stereo-consistency and temporal-consistency, without which viewers may experience 3D fatigue. There are several notable works on stereoscopic image super-resolution, but there is little research on stereo video {{super-resolution}}. In this paper, we propose a novel Transformer-based model for SVSR, namely \textit{Trans-SVSR}. \textit{Trans-SVSR} comprises two key novel components: a spatio-temporal convolutional self-attention layer and an optical flow-based feed-forward layer that discovers the correlation across different video frames and aligns the features. The parallax attention mechanism (PAM) that uses the cross-view information to consider the significant disparities is used to fuse the stereo views. Due to the lack of a benchmark dataset suitable for the SVSR task, we collected a new stereoscopic video dataset, SVSR-Set, containing $71$ full high-definition (HD) stereo videos captured using a professional stereo camera. Extensive experiments on the collected dataset, along with two other datasets, demonstrate that the \textit{Trans-SVSR} can achieve competitive performance compared to the state-of-the-art methods. Project code and additional results are available at \url{https://github.com/H-deep/Trans-SVSR/}.
  
\end{abstract}

\section{Introduction}
With augmented reality (AR) and virtual reality (VR) devices, dual-lens smartphones, and autonomous robots becoming widely accepted technologies worldwide, there is an increasing demand for various stereoscopic image/video processing techniques, including stereo editing, stereo inpainting, and stereo super-resolution. Stereo super-resolution is a fundamental low-level vision task that aims to enhance low-resolution (LR) stereo image/video spatial resolution by reconstructing it to the high-resolution (HR). A key challenge in stereo super-resolution is to preserve the stereo-consistency that may cause 3D fatigue to the viewers. While there are several prominent research on stereoscopic image super-resolution ({{Stereo ISR}}) \cite{wang2019learning, wang2021symmetric, zhu2021cross}, minor attention has been given to stereo video super-resolution (SVSR).  Compared to its image counterpart, SVSR presents an additional challenge of preserving temporal consistency. Therefore, the naive adoption of {{Stereo ISR}} to SVSR cannot achieve satisfactory performance.

Recently, utilization of deep learning, especially convolutional neural network (CNN) based methods, have shown great success for improving the {{Stereo ISR}} performance \cite{wang2019learning,ying2020stereo,song2020stereoscopic,wang2021symmetric}. They addressed the varying parallax, information incorporation, and occlusions and boundaries issues that exist in {{Stereo ISR}}. For example, Wang et al. \cite{wang2019learning} proposed the parallax attention module (PAM) that tackled the varying parallax problem in the parallax attention stereo SR network (PASSRnet). Ying et al. \cite{ying2020stereo} used several stereo attention modules (SAMs) with  pre-trained single image SR networks and addressed the information incorporation issue. Song et al. \cite{song2020stereoscopic} worked on solving the occlusion issue by designing a model for stereo consistency using disparity maps regressed with parallax attention maps. More recently, Wang et al. \cite{wang2021symmetric} used the intrinsic correlation within the stereo image pairs, and by using the symmetry cues, proposed a symmetric bi-directional PAM and an occlusion handling scheme to interact cross-view information. 

The direct extension of the stereoscopic image or conventional 2D video super-resolution methods to the stereoscopic video domain is challenging due to the need to maintain the disparity and temporal consistency simultaneously. Typically, relative object-camera motion between the neighboring frames in a (stereo) video is low. Therefore, the motion information between the consecutive frames can play a significant role in super-resolving the adjacent frames. Thus, the stereoscopic video super-resolution task can be divided into \textbf{1) modeling symmetry cues}  between two views, and \textbf{2) sequence modeling} between consecutive frames. The inherent correlation between pairs of stereo frames is used for symmetry modeling. The sequence modeling task can potentially be solved using recurrent neural networks (RNN), long short term memory (LSTM), and Transformers \cite{cao2021video}. Among these techniques, the more promising solution to tackle a sequence modeling task such as SVSR is the Transformers network \cite{cao2021video}, which is well-known for its capability in parallel computing and excellent performance in modeling the dependencies between the input sequences.

Transformer-based models for vision tasks such as Vision Transformers (ViT) \cite{dosovitskiy2020image} divide a video frame into small patches and extract the global relationships among the token embeddings which represents the patches, where local information is not given much attention \cite{li2021localvit}. These models cannot be directly applied for SVSR, in which the local and texture information is essential. Furthermore, temporal information and consistency, which are equally crucial in the SVSR task, cannot be solved by ViT. This paper proposes a novel Transformer-based model that can integrate the spatio-temporal information from the stereo views while maintaining both stereo- and temporal consistency. Specifically, after compensating the motion of previous and next frames, a Transformer network is applied to both the left and right views. We then use a CNN-based module to extract features and a modified PAM \cite{wang2019learning} module to fuse the features from the stereo views. Finally, the output is up-sampled, and a convolutional layer generates the super-resolved target frames.
The main contributions in this paper are summarized as follows: 

\begin{itemize}
    \item A new model, \textit{Trans-SVSR}, is proposed for the SVSR task. We designed a novel Transformer network, and the related parts of the model to make the proposed model suitable for the SVSR task.   {{Furthermore, the PAM module is modified and aligned to the SVSR.}}
 
    \item A novel optical flow-based feed-forward layer in our Transformer model that spatially align input features, by considering the correlations between all frames. 

    \item A new dataset, namely \textit{SVSR-Set}, is collected for the SVSR task. It contains $71$ high-resolution stereo videos in different indoor and outdoor settings, and is the largest dataset for the SVSR task. 

    \item Performance comparison 
    against several {{2D Video SR and}} {{Stereo ISR}} methods, re-implemented for SVSR on \textit{SVSR-Set} and two other datasets, demonstrates that Trans-SVR achieves state-of-the-art performance for the SVSR task.
\end{itemize}

\section{Related Works}
\label{Related Works}



{{ 
Recently, 2D video super-resolution (2D-VSR) is receiving increasing attention. As opposed to 2D image super-resolution (2D-ISR), 2D-VSR is more challenging since it entails combining data from numerous closely related but mismatched frames in video frames. There are two types of 2D-VSR techniques: sliding-window and recurrent methods. Previous approaches such as ToFlow \cite{xue2019video} predicts the flow across frames, followed by a warping process. Recent techniques use a more indirect approach. To align distinct frames at the feature level, SOF-VSR \cite{wang2020deep} uses deformable convolutions (DCNs) \cite{dai2017guodong}.To provide a smooth data flow and the preservation of texture features over extended periods, RRN \cite{isobe2020revisiting} uses a residual mapping across layers with skip connections. BasicVSR and IconVSR \cite{chan2021basicvsr} utilised 
essential functions, e.g. \textit{propagation}, \textit{alignment}, \textit{aggregation}, and \textit{upsampling}, with efficient designs, and showed that their method can achieve good efficiency and accuracy. 
}}

Traditional {{Stereo ISR}} methods mainly focused on estimating the disparity information and using this info for spatial resolution enhancement \cite{bhavsar2008resolution,bhavsar2010resolution}. Newer {{Stereo ISR}} methods exploit the cross-view information alongside the features from each mono-view image. For example, Jeon et al. \cite{jeon2018enhancing} did not calculate the disparity and used stereo images for super-resolution. They proposed a model to learn a parallax prior by training two networks and fusing the other view's spatial information by combining the left and shifted versions of the right images. Wang et al. \cite{wang2019learning} proposed a parallax-attention stereo super-resolution network (PASSRnet) to capture the stereo correspondence with a global receptive field along the epipolar line. In another study, Song et al. \cite{song2020stereoscopic} proposed a self and parallax attention mechanism (SPAM) that uses mono and cross-view information for {{Stereo ISR}}. They also developed a network and efficient loss functions to maintain stereo consistency. Xu et al. \cite{xu2021deep} used bilateral grid processing into a CNN network and proposed a bilateral stereo super-resolution network (BSSRnet) for {{Stereo ISR}}. The main idea is borrowed from image restoration. More recently, Wang et al. \cite{wang2021symmetric} enhanced the {{Stereo ISR}} by proposing the symmetric bi-directional parallax attention module (biPAM). Moreover, they recommended a scheme for inline occlusion handling to interact with cross-view information efficiently.

\begin{figure*}
\begin{center}
   \includegraphics[width=0.93\linewidth]{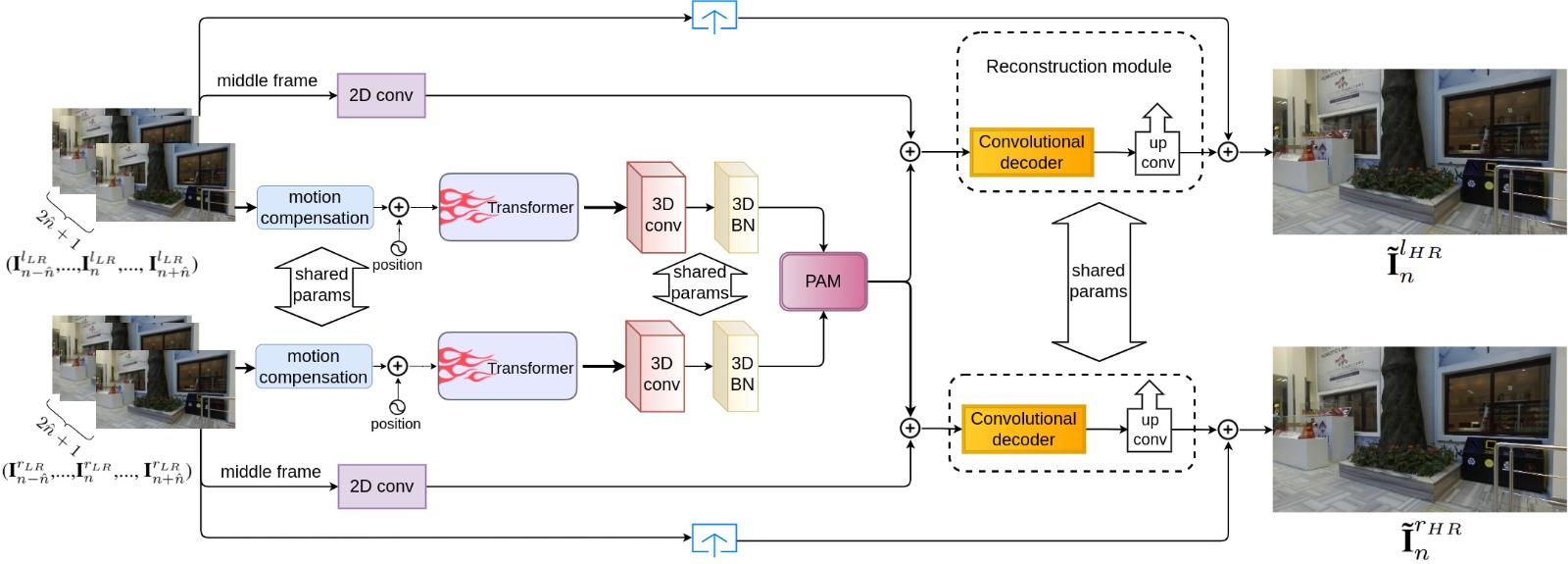}
\end{center}
   \caption{The architecture of the proposed model. The input is \(2\hat{n}+1\) left and right frames. The \(\hat{n}\) is the middle frame that super-resolves to the target frame. First, the neighboring frames are compensated for the motion computation. After position encoding, the frames are input to the Transformer with spatio-temporal architecture. Following the feature extraction using convolutional layers, PAM is used to fuse the left and right features. After feature extraction, the features are up-scaled. The super-resolved middle frames are the outputs.}
\label{fig:model}
\end{figure*}



Very recently, SVSRNet \cite{xu2021stereo} employed the view-temporal correlations for performing the SVSR task. They designed an attention module to combine the LR information from time dimension and stereo views to create HR stereo videos. Then, a fusion module is devised to fuse the information in the time dimension. They proposed a temporal and stereo views consistency loss function to enforce the consistency constraint of super-resolved stereo videos. {{They also developed a view-temporal attention mechanism for fusing the left and right view features. The PAM module is adopted to exploit the cross-view information further.}}

\section{Proposed Model}
\label{Proposedmethod}

The architecture of the proposed method is shown in Figure \ref{fig:model}. Given a batch of the consecutive stereo video frames, with the target frame to be super-resolved denoted as middle frame, we first estimate the motion between the middle (both left and right) frames and their corresponding neighbouring frames. Then, the neighbouring frames are warped to the middle frames. The middle frames and the frames compensated for motion are then passed to the Transformer for feature extraction. Next, the extracted features are passed to a 3D convolutional block to further extract more localized features. The extracted left and right features are then input to the PAM module \cite{wang2019learning} for fusing the features from the left and right middle frame pairs. The PAM features are then concatenated with extracted features of the middle frames and passed to a convolutional block. {{The convolutional decoder block consists of the consecutive 2D convolutions which is used for the creation of the super-resolved frames, and we name it as the reconstruction module in \ref{fig:model}. This module includes $8$ consecutive 2D convolutional layers, all with the kernel size of 3. The number of filters for each convolution layer are $256$, $512$, $1024$, $1024$, $512$, $256$, $128$, $3$, respectively.}} Finally, the features are up-scaled using up-convolutions and added with the up-scaled version of the middle frames, creating the super-resolved frames. We provide detailed description for each module in the following sections.

\noindent \textbf{Notations.} Let \(\textbf{I}_n^{l_{LR}}\) and \(\textbf{I}_n^{r_{LR}}\) be the n-\textit{th} low-resolution frames from the left and right stereo video \(\textbf{V}^{l_{LR}}\) and \(\textbf{V}^{r_{LR}}\), respectively. Our model's inputs are \(\textbf{I}_n^{l_{LR}}\) and \(\textbf{I}_n^{r_{LR}}\), and its aim is to create the high-resolution version of them as \(\textbf{\~{I}}_n^{l_{HR}}\) and \(\textbf{\~{I}}_n^{r_{HR}}\). 
For each view, we select the middle frames \(\textbf{I}_n^{l_{LR}}\) and \(\textbf{I}_n^{r_{LR}}\) from the consecutive frames as the frames aiming to be super-resolved. In addition, \(\hat{n}\) previous and next frames (\(\textbf{I}_{n-\hat{n}}^{l_{LR}}\),...,\(\textbf{I}_n^{l_{LR}}\),..., \(\textbf{I}_{n+\hat{n}}^{l_{LR}}\)) and (\(\textbf{I}_{n-\hat{n}}^{r_{LR}}\),...,\(\textbf{I}_n^{r_{LR}}\),..., \(\textbf{I}_{n+\hat{n}}^{r_{LR}}\)) are also selected as inputs to the model.

\begin{figure*}
\begin{center}
   \includegraphics[width=0.95\linewidth]{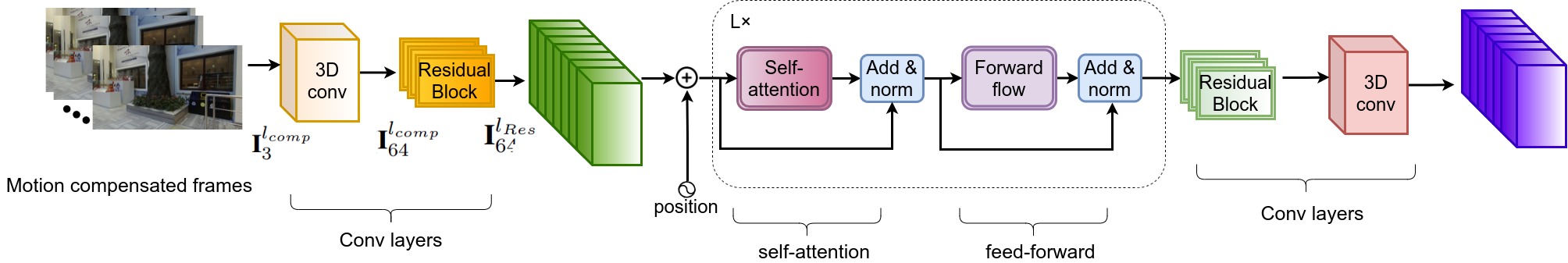}
\end{center}
   \caption{\small The high-level design architecture of the Transformer. Convolutional layers are used to extract features from the motion-compensated frames. After position encoding, the self-attention and feed-forward optical flows modules, both with residual connections to the add and normalization blocks, between them are applied. The final convolutional blocks provide the output features of the Transformer.}
\label{fig:tran}
\end{figure*}

\subsection{Motion Compensation}
{{Since Spatial Pyramid Network (SPyNet) \cite{ranjan2017optical} is used as baseline in optical flow computation,}} to warp each neighbouring frame (\(\textbf{I}_{n-\hat{n}}^{l_{LR}}\),..., \(\textbf{I}_{n+\hat{n}}^{l_{LR}}\)) and (\(\textbf{I}_{n-\hat{n}}^{r_{LR}}\),..., \(\textbf{I}_{n+\hat{n}}^{r_{LR}}\)) to the middle frame \(\textbf{I}_n^{l_{LR}}\) and \(\textbf{I}_n^{r_{LR}}\) by calculating the motion between the middle frame and each of the neighbouring frames. The warped frames provide different representations of the target frame. Let \textit{F} be the current optical flow field. The residual flow at each k-\textit{th} pyramid level \(\textit{f}_k\) for left frame is defined as: 
\begin{equation}\
f_k = Conv_k(\textbf{I}_{1,k}^{l_{LR}},warp(\textbf{I}_{2,k}^{l_{LR}},u(F_{k-1})),u(F_{k-1}))
\end{equation}
\noindent where \(\textit{u}\) is up-sampling operator, \(\textit{Conv}_k\) is the \(\textit{k}_{th}\) Convnet module, and \textit{warp} is the warping operator. The Convnet \(\textit{Conv}_k\) calculates the residual flow \(\textit{f}_k\) using the up-sampled flow from the previous level, \(\textit{F}_{k-1}\), and the frames \(\textbf{I}_{1,k}^{l_{LR}}\) and \(\textbf{I}_{2,k}^{l_{LR}}\) at level \textit{k}. The neighbouring frame \(\textbf{I}_{2,k}^{l_{LR}}\) is warped using the flow as \(warp(\textbf{I}_{2,k}^{l_{LR}},u(F_{k-1}))\). Finally, the flow at the k-\textit{th} level \(\textit{F}_k\) is:
\begin{equation}\
F_k = u(F_{k-1}) + f_k
\end{equation}

The process starts with the down-sampled frames from the top level of the coarse-to-fine pyramid by calculating flow \({f}_0\), at the top pyramid level. Then, we apply up-sampling to the resulting flow \(u({f}_0)\). Convolution, \(Conv_1\), is applied to the resulted up-sampled flow and \{\(\textbf{I}_{1,1}^{l_{LR}},warp(\textbf{I}_{2,1}^{l_{LR}},u(F_{0}))\)\} and \({f}_1\) is computed. This process is repeated for all pyramid levels. The left and right models shared the training parameters. We initialized the model with pre-trained weights from SPyNet and fine-tune the weights during the training.

\subsection{Transformer Architecture}
Our proposed Transformer block mainly includes a convolutional encoder, an attention layer, a feed-forward layer with skip connection, and a convolutional decoder. The high-level architecture of the Transformer is shown in Figure \ref{fig:tran}. Firstly, a 3D convolutional layer is applied to the input frame-batches to transfer the $3$-channel frames which has been compensated for motion (\(\textbf{I}_3^{l_{comp}}\) and \(\textbf{I}_3^{r_{comp}}\)) to $64$-channel features (\(\textbf{I}_{64}^{l_{comp}}\) and \(\textbf{I}_{64}^{r_{comp}}\)). With this operation, the number of attention heads in the Transformer could be increased \cite{zhou2021refiner}. Then, the residual blocks extract initial features from the input features (\(\textbf{I}_{64}^{l_{Res}}\) and \(\textbf{I}_{64}^{r_{Res}}\)). The encoder of our Transformer converts the features to a sequence of continuous representations. Self-attention and Feed-forward optical flows modules, with residual connections using the add and normalization blocks between them, are applied next. These layers are repeated \textit{L} times as shown in Figure \ref{fig:tran}. Finally, another residual block followed by a 3D convolutional layer is applied to the output representations. In the following sub-sections, the architecture of the sub-blocks of the transformer is explained.

\begin{figure}
\begin{center}
   \includegraphics[width=1.0\linewidth]{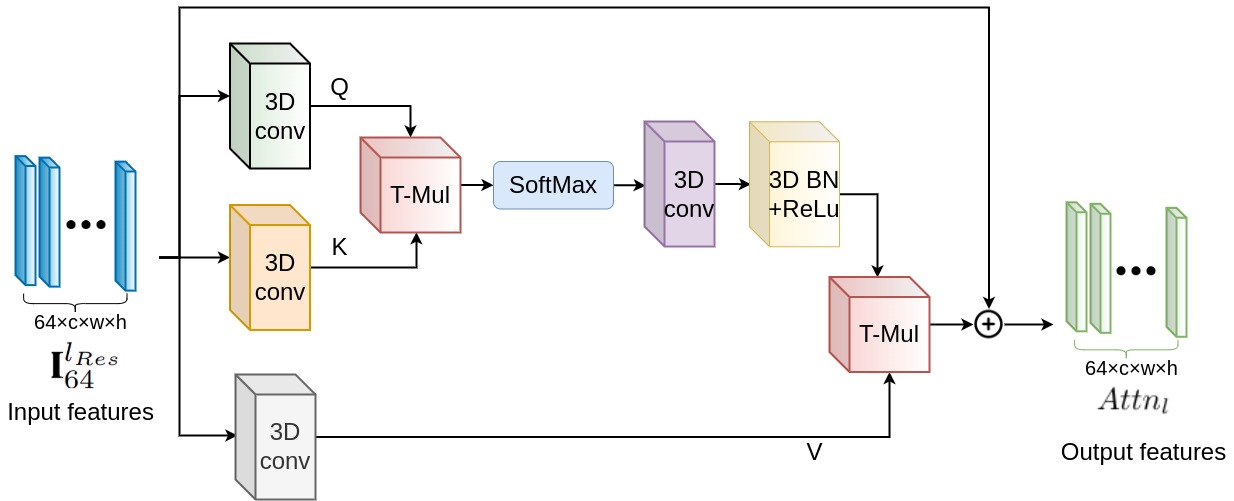}
\end{center}
   \caption{{{The architecture of the spatial-temporal convolutional self-attention module. Tensors Q, K, and V are created by passing the input features to a 3D convolutional block. Then, multiplication between Q and K, SoftMax operation, 3D convolution, and batch normalization create features to be further multiplied with V. Finally, the resulting features are added to the input features, and the output features are created.}}}
\label{fig:att}
\end{figure}

\textbf{Convolutional Self-attention.}
The architecture of the self-attention layer is shown in Figure \ref{fig:att}. Firstly we create the Query ({Q}), Key ({K}), and Value ({V}) tensors. With applying a 3D convolution to the input feature maps ((\(\textbf{I}_{64}^{l_{Res}}\) and \(\textbf{I}_{64}^{r_{Res}}\))), we create {Q} (\({Q}_{64}^{l}\) and \({Q}_{64}^{r}\)) and {K} tensors (\({K}_{64}^{l}\) and \({K}_{64}^{r}\)) in order to extract the spatio-temporal features of each input feature. $64$ filters with kernel size of 3$\times$3$\times$3 and padding of $1$ are used for all three convolution layers. The {Q}, {K}, and {V} for the left and right channels are as: 
\begin{equation}\
\begin{array}{l}
Q_l = Conv_{3D}(K_1,\textbf{I}_{64}^{l_{Res}})\\
K_l = Conv_{3D}(K_2,\textbf{I}_{64}^{l_{Res}})\\
V_l = Conv_{3D}(K_3,\textbf{I}_{64}^{l_{Res}})\\
\end{array}
\end{equation}
\begin{equation}\
\begin{array}{r}
Q_r = Conv_{3D}(K_1,\textbf{I}_{64}^{r_{Res}})\\
K_r = Conv_{3D}(K_2,\textbf{I}_{64}^{r_{Res}})\\
V_r = Conv_{3D}(K_3,\textbf{I}_{64}^{r_{Res}})\\
\end{array}
\end{equation}
\noindent where \({K}_{1}\), \({K}_{2}\), and \({K}_{3}\) are three independent convolutional kernels. Then, we calculate the similarity matrix using the tensor multiplication (\textit{TP}) and SoftMax operators: 
\begin{equation}\
\begin{array}{l}
QK_l = SoftMax(\textit{TP}(Q_l^T,K_l))\\
QK_r = SoftMax(\textit{TP}(Q_r^T,K_r))
\end{array}
\end{equation}

\noindent Following that, we feed the output features to a 3D convolutional layer with $64$ filters and kernel size of 3$\times$3$\times$3 with padding and stride of $1$, and a 3D batch normalization layer, and a ReLU activation function. The output features are then multiplied by {K} tensor and added with the input features to provide the output features of the attention layer:
{{
\begin{equation}
\begin{array}{l}
Attn_l = \textbf{I}_{64}^{l_{Res}} + \textit{TP}(QK_l,V_l)\\
Attn_r = \textbf{I}_{64}^{r_{Res}} + \textit{TP}(QK_r,V_r)
\end{array}
\end{equation}
}}

\textbf{Spatial-temporal positional encoding.}
The original Transformer architecture \cite{vaswani2017attention} is invariant to the permutation, but in super-resolution, {{the exact position information is important \cite{cao2021video}}}. We {{use} the positional encoding in \cite{wang2021end} to encode the 3D positional information of a video and add it along with the input to the attention block}. 
For each dimension coordinates, we use d/3 sinus and cosinus functions with different frequencies for the left and right Transformers as follows:





\({PE}_{l}(pos_l,i) = \)
$\left\{ 
  \begin{array}{ c l }
    sin(pos_l . w_k) & \quad \textrm{for i=2k,}  \\
    cos(pos_l . w_k) & \quad \textrm{for i=2k+1;}
  \end{array}
\right.$

\({PE}_{r}(pos_r,i) = \)
$\left\{ 
  \begin{array}{ c l }
    sin(pos_r . w_k) & \quad \textrm{for i=2k,}  \\
    cos(pos_r . w_k) & \quad \textrm{for i=2k+1;}
  \end{array}
\right.$

\noindent where {{k is the corresponding dimension. Specifically, each dimension of the positional encoding consistent with a sinus function. It will allow the model to easily learn to attend by
relative positions.}} \({pos}_{l}\) and \({pos}_{r}\) are the position in the corresponding dimension for left and right Transformers, respectively and \({w}_{k} = 1/10000^{2k/({d/3})}\) \cite{wang2021end}. Also, d is the channel dimension size and must be divisible by 3.

\begin{figure}
\begin{center}
   \includegraphics[width=1.0\linewidth]{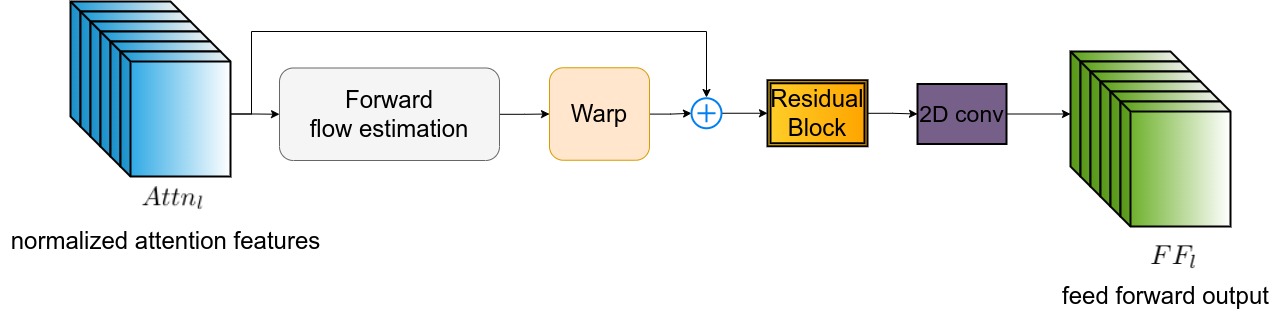}
\end{center}
   \caption{The optical flow-based feed-forward layer.} 
\label{fig:FF}
\end{figure}

\textbf{Flow-based Feed-Forward.} 
The traditional fully connected feed-forward layer includes two linear layers and is applied to each token identically. With this design, the fully connected feed-forward layer may not use the correlations between tokens related to the neighbouring frames. We proposed an optical flow-based method to spatially align the input features, taking into consideration the correlations between the input frames. The architecture of our flow-based feed-forward layer is shown in Figure \ref{fig:FF}. The input feature maps from the self-attention layer namely \({Attn}_{l}\) {{and \({Attn}_{r}\) are used as input to this module. Firstly, we calculate the optical flow between frame number $n$ (the middle frame) and frame number $m$ (where $m=1,...,5$) as}} \({flow}_{l}\) and \({flow}_{r}\):

{{
\({flow}_{l}(m,n) = \)
$\left\{ 
  \begin{array}{ c l }
    \left[0\right]_{W \times H}  & \quad \textrm{for m=n,}  \\
    spy(\textbf{I}_n^{l_{LR}}, \textbf{I}_m^{l_{LR}}) & \quad \textrm{for m$\not=$n;}
  \end{array}
\right.$

\({flow}_{r}(m,n) = \)
$\left\{ 
  \begin{array}{ c l }
    \left[0\right]_{W \times H}  & \quad \textrm{for m=n,}  \\
    spy(\textbf{I}_n^{r_{LR}}, \textbf{I}_m^{r_{LR}}) & \quad \textrm{for m$\not=$n;}
  \end{array}
\right.$
 
\noindent Next,}} we warp the input features along the forward direction{{, and concatenate (\textit{cat}) them with the input feature maps from the self-attention layer:
\begin{equation}\
\begin{array}{l}
FF_l = cat({Attn}_{l}, warp({Attn}_{l},{flow}_{l})) \\
FF_r = cat({Attn}_{r}, warp({Attn}_{r},{flow}_{r}))
\end{array}
\end{equation}
}}
{{
Then, we fuse the \({FF}_l\) and \({FF}_r\) with \({Attn}_{l}\) and \({Attn}_{r}\).
We propose a convolutional forward layer to establish the relationship between consecutive frames. Specifically, we use residual blocks and a 3D convolution layer with 1$\times$1$\times$1 kernel size, stride $1$, and zero padding, followed by the LeakyReLU activation function, to create the output features of this layer. The output feature size is $64$.
The fully connected feed-forward layer is defined as the following:

\begin{equation}\
\begin{array}{l}
FF_l^{o}({Attn}_{l}) = conv(LN( {Attn}_{l} + Res([{Attn}_{l},FF_l]))) \\
FF_r^{o}({Attn}_{r}) = conv(LN( {Attn}_{r} + Res([{Attn}_{r},FF_r]))) \\

\end{array}
\end{equation}

\noindent where $LN$, $conv$, $Res$ denote the layer normalization, convolution operation, and residual block, respectively. }}

{{

{{

\subsection{Modified PAM Architecture}
\label{Modified PAM}
Wang et al. \cite{wang2019learning} presented the parallax attention mechanism to estimate global matching in stereo images based on self-attention techniques \cite{zhang2019self,fu2019dual}. The left and right image pair's features can be efficiently merged using PAM.
Figure \ref{Fig20} depicts the structure of the redesigned PAM. A $1\times1$ layer receives the previous layer's output. The features are then sent to a SoftMax block to construct the attention maps M\textsubscript{R to L} and M\textsubscript{L to R}, which are created using batch-wised matrix multiplication. Next, the sum of features is combined with previous right features at all disparity levels.

To create additional features appropriate for deblurring, three 2D CNN layers are utilized, each followed by a ReLU activation function and a batch normalization layer. The first two layers, \textit{conv1} and \textit{conv2}, both consist of $128$ filters, but with different kernel size of 5$\times$5 and 3$\times$3 for  \textit{conv1} and \textit{conv2}{{,}} respectively. A dropout {{with}} rate of 0.5 is applied to conv2 layer, where random neurons' activity levels are forced to zero. The third layer \textit{conv3} comprises $64$ 3$\times$3 filters. {{To reduce the complexity of the whole model, we did not include the valid mask generation and the fusion parts in the original PAM.}}

}}

\begin{figure}
\centering
\includegraphics[width=1.0\linewidth]{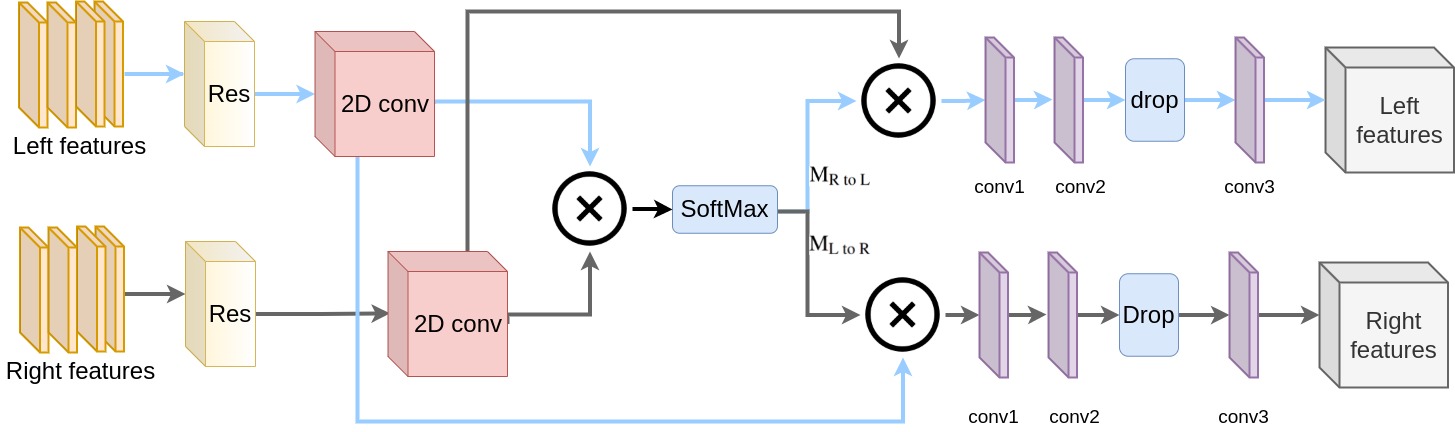}
\caption{
\label{Fig20}
\small 
The modified PAM architecture.}
\end{figure}

\subsection{Loss Functions}
We use three loss functions to train \textit{Trans-SVSR} on \textit{SVSR-Set}.
The first loss function is the mean absolute error (MAE) that calculates the average of the absolute differences between the HR and super-resolved frames. It is formulated as the average MAE of the left and right views:

\begin{equation}\
mae_{loss} = (mae_l + mae_r)/2
\end{equation}

\noindent where \(mae_l\) and \(mae_r\) are the MAE loss between HR and its corresponding super-resolved left and right frames, respectively. We also used photometric (\(photo_{loss}\)) and cycle (\(cycle_{loss}\)) losses \cite{wang2019learning} as additional loss functions. The total loss function is the combination of these three losses:

\begin{equation}\
loss = mae_{loss} + \lambda (photo_{loss} + cycle_{loss})
\end{equation}
\noindent where \(\lambda\) is the regularization term, empirically set as $0.01$.
}}

\section{Dataset Collection}
\label{DatasetandTraining}

For {{{{Stereo ISR}}}}, there are several datasets such as KITTI 2012 \cite{geiger2012we}, KITTI 2015 \cite{menze2015object}, Middlebury \cite{scharstein2014high} and Flickr1024 \cite{wang2019flickr1024}. These datasets mostly contain stereo images that may not be useful for the SVSR task. In the following sub-sections, we will discuss the existing datasets for SVSR, along with its limitation, followed by the detailed description of SVSR-Set, the new dataset we collected for the SVSR task.

\subsection{Existing Datasets}
KITTI 2012 \cite{geiger2012we} and KITTI 2015 \cite{menze2015object} datasets contain frames of videos with limited consecutive frames. Both datasets contain about $200$ stereo videos with limited frames. The time difference between the consecutive frames is significantly high, making them unsuitable for stereo video-related applications. On the other hand, SceneFlow \cite{mayer2016large} provides a dataset containing $2,265$ training and $437$ testing stereo videos. However, this dataset is synthetic. 

LFOVIAS3DPh2 \cite{appina2019study} (LFO3D) and NAMA\-3DS1-COSPAD1 \cite{urvoy2012nama3ds1} (NAMA3D) datasets are widely used for stereoscopic video quality assessment.
However, since these datasets are created originally for quality assessment purpose, many videos are distorted and of low quality. Notably, only 10 and 12 videos from Nama3D and LFO3D are in full HD resolution and suitable to be used for the SVSR task. 
The LFO3D videos are chosen from the RMIT3DV \cite{cheng2012rmit3dv} video dataset that includes symmetrically and asymmetrically distorted videos. All the videos in {RMIT3DV} dataset are captured using a Panasonic AG-3DA1 camera with full HD 1920$\times$1080 resolution. The time duration of all videos in the dataset is constant, which is only $10$ seconds. In summary, these datasets are too small for training a good model for the SVSR task. 

\begin{table}[b] 
\centering
\caption{Illustration of the \textit{SVSR-Set}. Stimuli type, light conditions (day/night), motion strength (low/high), indoor/outdoor, and the number of stereo videos for each category is shown in this table.}
\scalebox{0.75}{
\begin{tabular}{| l |@{\hspace{10pt}} *{5}{c|} |}
	\hline 
    {\textbf{Stimuli Type}} & {\textbf{\# videos}} & {\textbf{Light}} & {\textbf{Motion}}& {\textbf{Setting}} \\
	\hline
    people,tree & 5 & day & high & Outdoor\\
    \hline
    people,tree,car,motor & 12 & day & high & Outdoor\\
    \hline
    people,tree,car,motor & 4 & night & high & Outdoor\\
    \hline
    people,train & 2 & day & high & Outdoor\\
    \hline
    people,ship,water & 9 & day & low & Outdoor\\
    \hline
    bird,dog,grass & 4 & day & low & Outdoor\\
    \hline
    grass,motor & 3 & day & high & Outdoor\\
	\hline 
    water,bird & 8 & day & high & Outdoor\\
	\hline 
    people & 5 & night & low & Outdoor\\
	\hline 
    toy & 6 & day & high & Indoor\\
	\hline 
    game & 2 & day & high & Indoor\\
	\hline 
    flower & 4 & day & high & Outdoor\\
	\hline 
    people & 7 & day & low & Indoor\\
	\hline 
\end{tabular}}
\label{tablesr}
\end{table}

\begin{table}[htb]
\centering
\caption{Comparison of the three datasets used in the experiments. Number of the videos refer to the reference stereo videos.}
\scalebox{0.75}{
\begin{tabular}{| l |@{\hspace{8pt}} *{4}{c|} |}
	\hline 
    {\textbf{Dataset}} & {\textit{SVSR-Set}} & {NAMA3D \cite{urvoy2012nama3ds1}} & {LFO3D \cite{appina2019study}} \\
	\hline
    Video No. & 71 & 10 & 12\\
    \hline
    FPS & 30 & 25 & 30\\
    \hline
    Duration & 20 & - & 10\\
    \hline
    Resolution & full-HD & full-HD & full-HD\\
    \hline
\end{tabular}}
\label{setcomp}
\end{table}

\subsection{SVSR-Set Dataset}
Due to the lack of an existing real-world dataset that is sufficiently large for training a good deep neural network model for the SVSR task, we developed \textit{SVSR-Set}, a new real-world dataset, which can be used for training and evaluating the SVSR methods in the future. Compared to 2D video datasets, the creation of stereo video datasets is more challenging and time-consuming.  This dataset contains $71$ stereo videos, collected using a professional ZED 2 stereoscopic camera \cite{WinNT}. Each stereo video clip is recorded as a full high-definition (HD) (1080$\times$1920) video and contains $20$ seconds video capture at $30$ frames per second (FPS). These videos are available in .svo and .avi containers format. The videos are recorded in various settings, including indoor and outdoor, low and high motion, low and high illumination, etc. We also provide the disparity ground truth for two consecutive frames. The\textit{ SVSR-Set} is made publicly available to the research community at \url{http://shorturl.at/mpwGX}.

\textbf{Calibration.} 
We calibrated the camera to ensure that possible slight shifts in the camera's internal parts would not make the dataset unusable. The calibration file was created one time and used for the whole recording process. The calibration file includes details about the exact location of the left and right cameras and their optical properties. First, we turned the lights off and closed window blinds that may cause reflections on the screen and made the calibration process longer. During the calibration process, a grid window and a red dot will appear in the middle of the monitor screen. When the camera is put in front of the monitor, a blue dot will appear. The aim of the calibration is to match the blue dot with the red dot. This process is repeated several times because the dots has motion and hence, may change its location.

\textbf{Dataset collection.} The moving objects contain people, ships, flowers, and other objects from public places. The video attributes; stimuli type, light conditions (day/night), motion strength (low/high), indoor/outdoor, and the number of stereo videos, are shown in Table \ref{tablesr}. As seen, most of the stereo videos contain "people" stimuli. $12$ videos contain the "people, tree, car, motor" objects, recorded during outdoor daytime settings, and the high motion strength, $15$ videos are recorded in the indoor settings, $9$ videos captured at night with low light conditions. 
Table \ref{setcomp} compares the attributes of \textit{SVSR-Set} dataset with the existing datasets.

\subsection{Implementation Details}
We trained our proposed \textit{Trans-SVSR} on the developed \textit{SVSR-Set}. We randomly split the stereoscopic videos in the SVSR-Set dataset to train and test sets. Our training set contains $58$ stereo videos, and the testing set has $13$ videos. We down-sampled the HR frames using the bicubic method to create LR images. For training, we cropped the LR frames and their corresponding HR frames to non-overlapping patches of size 32$\times$88. We conducted experiments on 6$\times$SR and 4$\times$SR. For 6$\times$SR and 4$\times$SR videos, the size of HR training frames are 192$\times$528 and 128$\times$352. We have created $519,390$ and $1,385,040$ training samples for 6$\times$SR and 4$\times$SR, respectively.

We used a computing system with the following specifications: i9-10850K CPU 3.60 GHz, 64GB memory, NVIDIA GeForce RTX 3090 GPU with $24$GB of GPU memory. The Adam optimizer \cite{kingma2014adam} with parameters \({\beta}_{1}\)=0.90 and \({\beta}_{2}\)=0.99, is used in our experiments. We used a batch size of $7$ and $1,385$k iterations to train the proposed model. The models are implemented with PyTorch 1.8.0 library. {{The learning rate is initialized to 2\textit{e-4}, and reduced by 50\% per epoch.
}}

\section{Results and Discussions}
\label{Results and Discussions}
Evaluation of our proposed method, \textit{Trans-SVSR} is performed on {{three stereo video datasets; \textit{SVSR-Set}, LFO3D \cite{appina2019study}, and NAMA3D \cite{urvoy2012nama3ds1}}}. We could not compare our method with the very recently proposed SVSRNet \cite{xu2021stereo}, which is also the only SVSR method, because of unavailability of open-source code.  Therefore, we compare our method with the recently published state-of-the-art {{Stereo ISR}} {{and 2D video SR}} methods. We re-implemented iPASSR \cite{wang2021symmetric}, PASSRnet \cite{wang2019learning}, SRRes+SAM \cite{ying2020stereo}, DFAM \cite{dan2021disparity} and its variants as well as {{two 2D video SR methods; SOF-VSR \cite{wang2020deep} and RRN}} to perform testing on the three datasets. 

\begin{table}[b]
\centering
\caption {Performance comparison of our proposed \textit{Trans-SVSR} and state-of-the-art {{{{Stereo ISR}}}} methods for 4$\times$SR videos on three datasets: \textit{SVSR-Set}, LFO3D \cite{appina2019study}, and NAMA3D \cite{urvoy2012nama3ds1}. 
The best results are in \textbf{Bold} and the second-best results are \underline{underlined}.} 
\scalebox{0.57}{
\begin{tabular}{llllllllll}
 \textbf{Dataset} & & \multicolumn{2}{c}{\textbf{SVSR-Set}} &  & \multicolumn{2}{c}{\textbf{NAMA3D \cite{urvoy2012nama3ds1}}}  & & \multicolumn{2}{c}{\textbf{LFO3D \cite{appina2019study}}} \\ 
\cline{3-4} \cline{6-7}  \cline{9-10}
      \textbf{{{Stereo ISR}} methods}  & {{N.Par}} & PSNR  & SSIM   &             
	&  PSNR  & SSIM  & & PSNR  & SSIM   \\    \hline

PASSRnet \cite{wang2019learning} &  1.41M & {28.9723} & 0.8957 & & 25.8708 & 0.8335 & & 22.4558 & 0.7047  \\
SRRes+SAM \cite{ying2020stereo} &  1.73M &  27.2863 & 0.8681 && 24.2226 & 0.7978 &&  19.4583  & 0.6736\\
SRResNet-DFAM \cite{dan2021disparity}  & 2.89M &  27.5388  & 0.8905 && 24.8192 & 0.8212 &&  21.0419  & 0.7389 \\
SRCNN-DFAM \cite{dan2021disparity} & 0.73M & {{27.7202}}   & {0.9008} && {24.5390} & {0.8344} &&  {21.2752}  & {0.7355}\\

VDSR-DFAM \cite{dan2021disparity} & 2.68M &  28.4919  & 0.8949 && 25.2385 & 0.8401 &&  22.1496  & {0.7435}\\
RCAN-DFAM \cite{dan2021disparity}& 16.9M &  {{29.0158 }}   & {{0.9013}} && { \underline{25.9539}} & {\underline{0.8442} } &&  {{22.5685}}  & {0.7427}\\
iPASSR \cite{wang2021symmetric}   & 1.42M &  28.1980  & 0.8913 && {24.7818} & {0.8195} &&  21.4525  & 0.6865\\
 \hline 
      \textbf{2D-VSR methods}  &  &   &    &             
	&    &   & &   &    \\    \hline
  SOF-VSR \cite{wang2020deep}  & 2.08M  & {28.7208}  & {0.9002} &    & 24.573  & 0.8401  &&  {22.5642} & {0.7414} \\
 RRN \cite{isobe2020revisiting} & 14.40M  & \underline{29.1454}  & \underline{0.9069} &    & 24.6100  & 0.8418  &&  \underline{22.7596} & \underline{0.7483} \\

 \hline 
      \textbf{SVSR methods}  &  &   &    &             
	&    &   & &   &    \\    \hline

 \textbf{Trans-SVSR} & 27.29M  & \textbf{31.9766}  & \textbf{0.9293} &    & \textbf{28.8424}  & \textbf{0.8674}  &&\textbf{25.5871}  & \textbf{0.7642} \\
 \hline 

\end{tabular}}
\label{table8}
\end{table}

\subsection{Quantitative Performance}
Quantitative evaluation of the proposed method is considered using two measures; namely, peak signal-to-noise ratio (PSNR) and structural similarity (SSIM) \cite{wang2004image} in RGB space. Table \ref{table8} compares results of our proposed \textit{Trans-SVSR} method with other methods on the \textit{SVSR-Set}, LFO3D \cite{appina2019study}, and NAMA3D \cite{urvoy2012nama3ds1} datasets for 4$\times$SR videos. As shown in this table, our proposed method achieves the state-of-the-art results compared to the {{{{Stereo ISR}} and 2D video SR}}-based methods on all three datasets. The PSNR of \textit{Trans-SVSR} on \textit{SVSR-Set} is{{ \textit{31.9766}, which is \textit{2.8312} dB }}better than the second-best performing RRN method, and is considered a significant improvement. The SSIM value for our method is {{\textit{0.9293}, which again, outperformed all methods in comparison.}} As can be seen from the results on LFO3D \cite{appina2019study} dataset, all methods obtained lower PSNR and SSIM on this dataset. It shows that this dataset is more challenging for all SR methods, largely due to the small dataset size. Interestingly, for this dataset, our method again achieves state-of-the-art performance compared to the other methods. {{On LFO3D \cite{appina2019study} dataset, compared to the second best-performing methods, PSNR and SSIM for our method are improved by $2.8275$ and $0.0159$ dB,}} respectively. PSNR and SSIM for our method on NAMA3D \cite{urvoy2012nama3ds1} dataset are $28.0543$ and $0.8473$, which are higher than all the {{{{Stereo ISR}}}}-based methods.

\textbf{Train and test on 6$\times$SR videos.} We also trained our model with \textit{L=20} on 6$\times$SR videos in \textit{SVSR-Set} training set, and tested on \textit{SVSR-Set} test set, NAMA3D \cite{urvoy2012nama3ds1}, and LFO3D \cite{appina2019study} datasets. Table \ref{table6} shows the testing results of our method on these three datasets.

\begin{table}[b]
\centering
\caption {Performance of the \textit{Trans-SVSR} for 6$\times$SR videos on three datasets. PSNR and SSIM are calculated based on the average between left and right HR and super-resolved stereo frame pairs.} 
\scalebox{0.67}{
\begin{tabular}{lllllllll}
 \textbf{Dataset} & \multicolumn{2}{c}{\textbf{SVSR-Set}} &  & \multicolumn{2}{c}{\textbf{NAMA3D \cite{urvoy2012nama3ds1}}}  & & \multicolumn{2}{c}{\textbf{LFO3D \cite{appina2019study}}} \\ 
\cline{2-3} \cline{5-6}  \cline{8-9}
        &  PSNR  & SSIM   &             
	&  PSNR  & SSIM  & & PSNR  & SSIM   \\    \hline
 \textbf{Trans-SVSR}   &  28.2775 & 0.8588   &  & 25.0481
    & 0.7684   &  &  22.3305   & 0.60187 \\
 \hline 
\end{tabular}}
\label{table6}
\end{table}

\begin{figure*}
\begin{center}
   \includegraphics[width=0.88\linewidth]{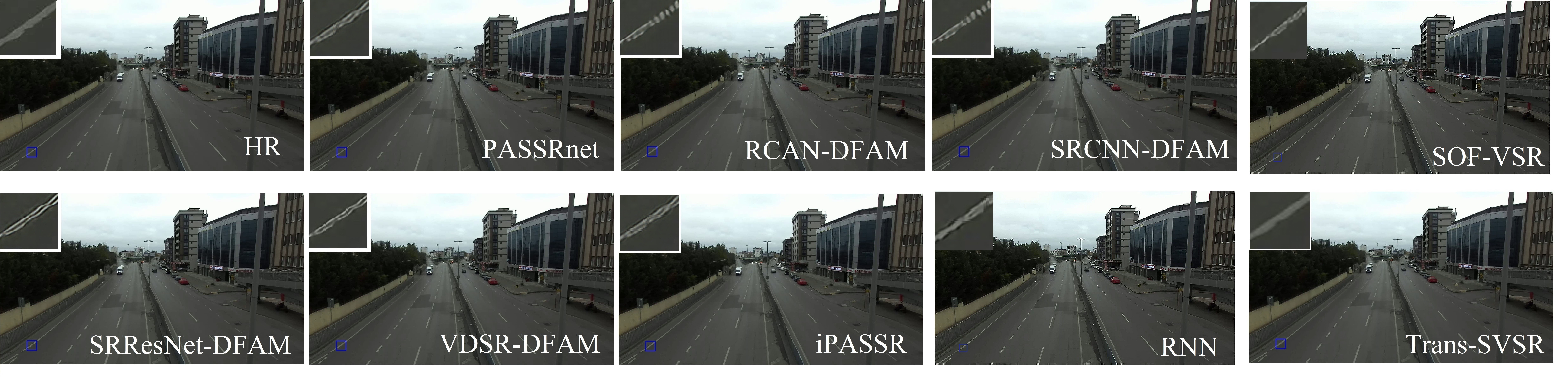}
\end{center}
   \caption{Qualitative results. One frame from \textit{SVSR-Set} dataset. We compared \textit{trans-SVSR} with PASSRnet \cite{wang2019learning}, iPASSR \cite{wang2021symmetric}, RCAN-DFAM, SRCNN-DFAM, SRResNet-DFAM, VDSR-DFAM \cite{dan2021disparity}, {{SOF-VSR \cite{wang2020deep}, and RRN}} \cite{isobe2020revisiting}.
   }
\label{fig:qua}
\end{figure*}

\textbf{Stereo consistency. }To measure the consistency between super-resolved frames of the output of our model and the reference stereo frames, we compute the end-point error (EPE) by calculating the Euclidean distance between the disparity \cite{hirschmuller2007stereo} of the super-resolved frames and the reference frames.
From the comparison of EPE with other methods in Table \ref{consis}, it can be observed that the proposed \textit{Trans-SVSR} better preserves the stereo disparity on \textit{SVSR-Set} dataset.

\begin{table}[htb]
\centering
\caption {{{Comparison of stereo consistency between our \textit{Trans-SVSR}, and the {{Stereo ISR}}-based methods on \textit{SVSR-Set} dataset.}}} 
\scalebox{0.7}{
\begin{tabular}{lllll}
Model    &   PASSRnet \cite{wang2019learning} &   RCAN-DFAM \cite{dan2021disparity}&   iPASSR \cite{wang2021symmetric}  &   \textbf{Trans-SVSR}   \\   
     \hline
  Avg EPE & 0.7175 & 0.6006 & 0.6212 & 0.5031 \\
 \hline 
\end{tabular}}
\label{consis}
\end{table}

\subsection{Qualitative Performance}
Qualitative results for 4$\times$SR are shown in Figure \ref{fig:qua}. From the results in the top rows, it can be observed that {{Stereo ISR}} methods tend to made the frame sharper, which indirectly resulted in more noise. Results of our method appear to be smoother, less disjointed and less halo effect as compared to some {{Stereo ISR}} results; i.e. RCAN-DFAM and SRCNN-DFAM. The reason could be that {{Stereo ISR}} methods generally use the spatial information of one view, or cross-view information for performing super-resolution, disregarding the temporal information. In contrast, our proposed method, \textit{Trans-SVR} also uses the temporal information from the neighboring frames, alongside the spatial information, and hence it can create smoother results with more details.

\subsection{Ablation Study} 

The ablation studies investigate the effect of removing different modules on the performance of Trans-SVSR.

\begin{table}
\centering
\caption {{{Performance comparison of \textit{Trans-SVSR}, with and without different contributing modules, on the \textit{SVSR-Set} dataset.}}}
\scalebox{0.8}{
\begin{tabular}{llll}
    \hline
    Model type    & {{N.Par}} & PSNR  & SSIM    \\   
    \hline
    {Trans-SVSR-WMF} & 27.29M  & {28.9715}  & {0.8619}  \\
    \hline
    {Trans-SVSR-WOT} & 23.47M  & {29.8914}  & {0.8959}  \\
    \hline 
    {Trans-SVSR-WOP} & 27.20M  & {30.6348}  & {0.9068} \\
    \hline 
    {Trans-SVSR-WOR} & 9.48M  & {30.3605}  & {0.9031} \\
    \hline 
    {Trans-SVSR-WSF}  & 27.29M &  30.3129 & 0.8989  \\
    \hline
    \textbf{Trans-SVSR} & 27.29M  & \textbf{31.9766}  & \textbf{0.9293} \\
    \hline 
\end{tabular}}
\label{ablation}
\end{table}

\noindent \textbf{Effect of the temporal frames.} 
To see the effect of the motion in the input frames on the performance of the proposed model, we trained \textit{Trans-SVSR} with just the middle stereo frames as the input (\textit{Trans-SVSR-WMF}). We repeated the middle frame $5$ times for the left and right input stream and fed it to the model. In this way, the motion information from the stereo video is removed. The first row of Table \ref{ablation} shows that without including the neighboring frames, the model performance decreases considerably. This study indicates that the adjacent frames and the motion between them are used effectively in \textit{Trans-SVSR}.

\noindent \textbf{Influence of the Transformer.} 
Results of removing the Transformer (\textit{Trans-SVSR-WOT}) is depicted in the second row of Table \ref{ablation}. From the results, we can observe that the Transformer greatly influences the model performance, whereby PSNR decreases by $2.0852$ dB when the Transformer is removed from both the left and right channels. Comparing with the results of other ablation study in Table \ref{ablation}, it is evident that Transformer plays an important role in boosting the performance of the proposed model.

\noindent \textbf{Effect of PAM.} The PAM module helps to fuse the left and right information and cross-view information when deblurring one view. By disabling the PAM module (\textit{Trans-SVSR-PAM}), the left and right channels will act like two different models, leading to decrease in performance as shown by results in the third row of Table \ref{ablation}. This result demonstrates that the cross-view information influences the model performance positively.

\noindent \textbf{Influence of the reconstruction module.} 
To investigate the absence of the reconstruction module to the performance of our model, we removed this module  from our network (\textit{Trans-SVSR-WOR}). Results in the fourth row of Table \ref{ablation} show that the performance drops considerably. This result shows that without the decoder block, the model lacks reconstruction capability.

\noindent \textbf{Impact of flow-based feed-forward layer.} 
To show the effectiveness of the flow-based feed-forward layer of the proposed Transformer, we conducted additional experiments. The fifth row of Table \ref{ablation} shows the comparison results of our model performance, which uses the flow-based feed-forward layer (\textit{Trans-SVSR-WSF}) with the standard feed-forward layer (\textit{Trans-SVSR}). As this table shows, on \textit{SVSR-Set} dataset, a decrease of $1.6161$ dB is resulted, which is a considerable difference.

\section{Conclusion and Future Works}
\label{Conclusion}
This paper proposed a novel stereoscopic video super-resolution framework {{based on a spatio-temporal Transformer network}}. We designed the self-attention and optical flow-based feed-forward layer to make the Transformer suitable for SVSR. In addition, we collected a new  \textit{SVSR-Set} dataset that can also be used for both the SVSR and {{Stereo ISR}} tasks. 
We trained our model on the \textit{SVSR-Set} dataset. Comparison with the state-of-the-art {{Stereo ISR}} methods on three datasets demonstrates that our method {{achieves the state-of-the-arts results.}} 
One limitation of the \textit{Trans-SVSR} method is the obviously larger number of the network parameters, as shown in Table \ref{table8}. However, this is probably reasonable since our method is proposed to address video super-resolution with an additional dimension as compared to the {{Stereo ISR}} task. The added temporal dimension unavoidably resulted in a more complex model.
In the future, we will design loss functions that can better reflect the quality difference between HR and super-resolved video frames. To reduce complexity, model pruning can be used to minimize computational and storage requirements for model inference \cite{chen2021structured}.

\section*{Acknowledgement}
\noindent This work is supported by the Scientific and Technological Research Council of Turkey (TUBITAK) 2232 Leading Researchers Program, Project No. 118C301.

{\small
\bibliographystyle{ieee_fullname}
\bibliography{main}
}

\end{document}